\documentclass[letterpaper, 10 pt, conference]{ieeeconf}  % Comment this line out if you need a4paper
\usepackage{cite}
\usepackage{amsmath,amssymb,amsfonts}
\usepackage{graphicx}
\usepackage{threeparttable}
\usepackage{booktabs}
\usepackage{makecell} % allow \\ in cells
\usepackage{multirow} % allow one cell with multiple rows
\usepackage{textcomp}
\usepackage[colorlinks,linkcolor=blue,citecolor=black]{hyperref}
\usepackage{algpseudocode}
\usepackage{algorithm}
\usepackage{xcolor}
\usepackage{amsmath}

\IEEEoverridecommandlockouts                              % This command is only needed if 
\title{\LARGE \bf
Preparation of Papers for IEEE Sponsored Conferences \& Symposia*
}

\title{\LARGE \bf TransForce: Transferable Force Prediction for \\ Vision-based Tactile Sensors with Sequential Image Translation  }
\author{Zhuo Chen$^{1}$, Ni Ou$^{1,2}$, Xuyang Zhang$^{1}$ and Shan Luo$^{1}$
\thanks{*This work was supported by the EPSRC project ``ViTac: Visual-Tactile Synergy for Handling Flexible Materials" (EP/T033517/2).}
\thanks{$^{1}$Zhuo Chen, Ni Ou, Xuyang Zhang and Shan Luo are with the Robot Perception Lab, Centre for Robotics Research, Department of Engineering, King's College London, London WC2R 2LS, United Kingdom. Emails: {\tt\small \{zhuo.7.chen, shan.luo\}@kcl.ac.uk}.}
\thanks{$^{2}$Ni Ou is with the State Key Laboratory of Intelligent Control and Decision of Complex Systems, Beijing Institute of Technology, Beijing, 100081, China.}
}

\begin{document}
\maketitle
\thispagestyle{empty}
\pagestyle{empty}

%%%%%%%%%%%%%%%%%%%%%%%%%%%%%%%%%%%%%%%%%%%%%%%%%%%%%%%%%%%%%%%%%%%%%%%%%%%%%%%%
\begin{abstract}

Vision-based tactile sensors (VBTSs) provide high-resolution tactile images crucial for robot in-hand manipulation. However, force sensing in VBTSs is underutilized due to the costly and time-intensive process of acquiring paired tactile images and force labels. In this study, we introduce a transferable force prediction model, TransForce, designed to leverage collected image-force paired data for new sensors under varying illumination colors and marker patterns while improving the accuracy of predicted forces, especially in the shear direction. Our model effectively achieves translation of tactile images from the source domain to the target domain, ensuring that the generated tactile images reflect the illumination colors and marker patterns of the new sensors while accurately aligning the elastomer deformation observed in existing sensors, which is beneficial to force prediction of new sensors. As such, a recurrent force prediction model trained with generated sequential tactile images and existing force labels is employed to estimate higher-accuracy forces for new sensors with lowest average errors of 0.69N (5.8\% in full work range) in $x$-axis, 0.70N (5.8\%) in $y$-axis, and 1.11N (6.9\%) in $z$-axis compared with models trained with single images. The experimental results also reveal that pure marker modality is more helpful than the RGB modality in improving the accuracy of force in the shear direction, while the RGB modality show better performance in the normal direction.

\end{abstract}

%%%%%%%%%%%%%%%%%%%%%%%%%%%%%%%%%%%%%%%%%%%%%%%%%%%%%%%%%%%%%%%%%%%%%%%%%%%%%%%%
\section{INTRODUCTION}\label{intro}

\begin{figure}[htbp]
\centering
\includegraphics[width=0.4\textwidth]{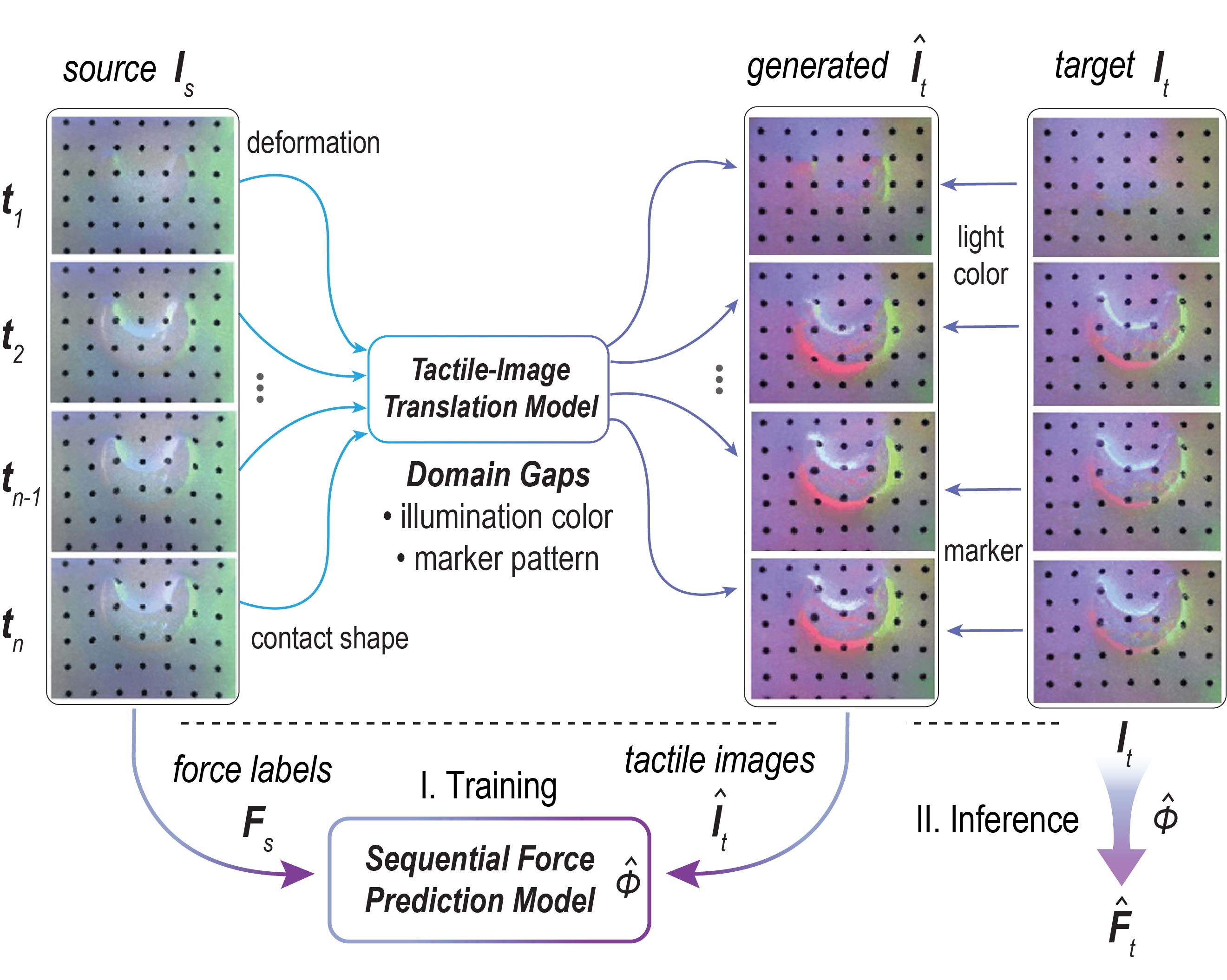}
\caption{Transferable force prediction model for VBTSs. The image translation model takes tactile images $\mathbf{I}_s$ from the source domain as inputs and generates tactile images $\hat{\mathbf{I}}_t$, which share similar illumination colors and marker patterns with the target domain $\mathbf{I}_t$ while aligning deformation and contact shape with the source domain $\mathbf{I}_s$. After (I) training with sequential tactile images $\hat{\mathbf{I}}_t$ and force labels $\mathbf{F}_s$, (II) the force prediction model $\hat{\phi}$ is able to infer forces $\mathbf{\hat{F}}_t$ with sequential tactile images $\mathbf{I}_t$.}
\label{fig1}
\end{figure}

Vision-based tactile sensors have now been widely used in robot manipulation \cite{she2021cable,liu2023gelsight,han2024learning}. By measuring high-resolution tactile images, robots equipped with VBTSs are endowed with the sense of touch to complete in-hand tasks with human-like dexterity, such as slip detection \cite{ou2024marker, lloyd2024pose} and dexterous manipulation \cite{suresh2023neural,yang2024anyrotate}. 

Recently, deep learning models \cite{insight,do2023densetact,lin20239dtact} are developed to map tactile images to force values without the aid of a physical model \cite{zhang2022tac3d,li2024biotactip}. Training these image-force mapping models requires a substantial amount of paired image-force data, along with costly calibration instruments like 6-DoF force/torque sensors. This labor-intensive data collection process must be repeated frequently due to wear and tear of soft elastomer and changes in sensor components, such as camera and LED. Moreover, structural variations among different VBTSs families, including bias of illumination colors and marker patterns, create significant domain gaps in visual features of collected tactile images. These domain gaps severely impact the transferability of trained force mapping models, posing challenges for the deployment of force sensing in unseen VBTSs.

Transfer learning \cite{zhuang2020comprehensive} has demonstrated strong performance in transferring learned models to mitigate domain gaps for new tasks; however, the collection of new paired image-label data remains necessary. Recently, domain adaptation regression (DAR) methods \cite{chen2024deep} have proven effective in learning a shared regressor to predict force across different domains by aligning the feature space of tactile images. While this approach excels in predicting normal forces, it struggles with high errors in shear force prediction, which needs improvement for real robotic applications. Therefore, there remains a demand for a transferable force prediction method with high-accuracy for VBTSs that can effectively predict both normal and shear forces.

In this study, we propose a novel transferable model, TransForce, to address the challenge of unsupervised adaptation in force prediction for VBTSs, with a particular focus on improving the accuracy of shear force. As illustrated in Fig. \ref{fig1}, TransForce translates the style of tactile images from a source domain, where force labels are available, into the target domain, representing new sensors. This process successfully preserves the sequential deformation information from the source domain while adapting image properties, such as illumination color, intensity, and marker pattern, to the target domain. Consequently, a recurrent force prediction model can be trained using the generated sequential tactile images and existing force labels, enabling accurate force estimation with tactile images from new sensors. Experimental results validate the effectiveness of TransForce in translation of tactile images across different domains, demonstrating low prediction errors in both normal and shear directions. Furthermore, we reveal that among different types of input tactile images, pure marker-based images are more useful in shear force prediction, while images with RGB information are superior for normal force prediction.

To the best of our knowledge, this is the first work to leverage a generative model combined with a recurrent neural network to tackle this regression challenge. The key contributions of this work are as follows:

\begin{enumerate}
    \item We propose a transferable model that utilizes image translation technique and sequential force prediction to adapt existing paired image-force data to new VBTSs.
    \item The model can predict high-accuracy forces within a wide range of -6 N to 6 N in the shear direction and -16 N to 0 N in the normal direction, tested across 18 types of seen and unseen indenters;
    \item Our experiments find that pure marker modality is superior in shear force prediction while RGB information is helpful for normal force prediction. 
\end{enumerate}

The rest of the paper is structured as follows: Section~\ref{sec:relatedworks} provides an overview of related works; Section~\ref{sec:methodology} introduces our methodology; Section~\ref{Sec:IV} details our data collection and implementation; Section~\ref{sec:experiment} analyses the experimental results. Finally, Section~\ref{sec:conclusion} summarises the work.

\begin{figure*}[t]
\centering
\includegraphics[width=0.75\textwidth]{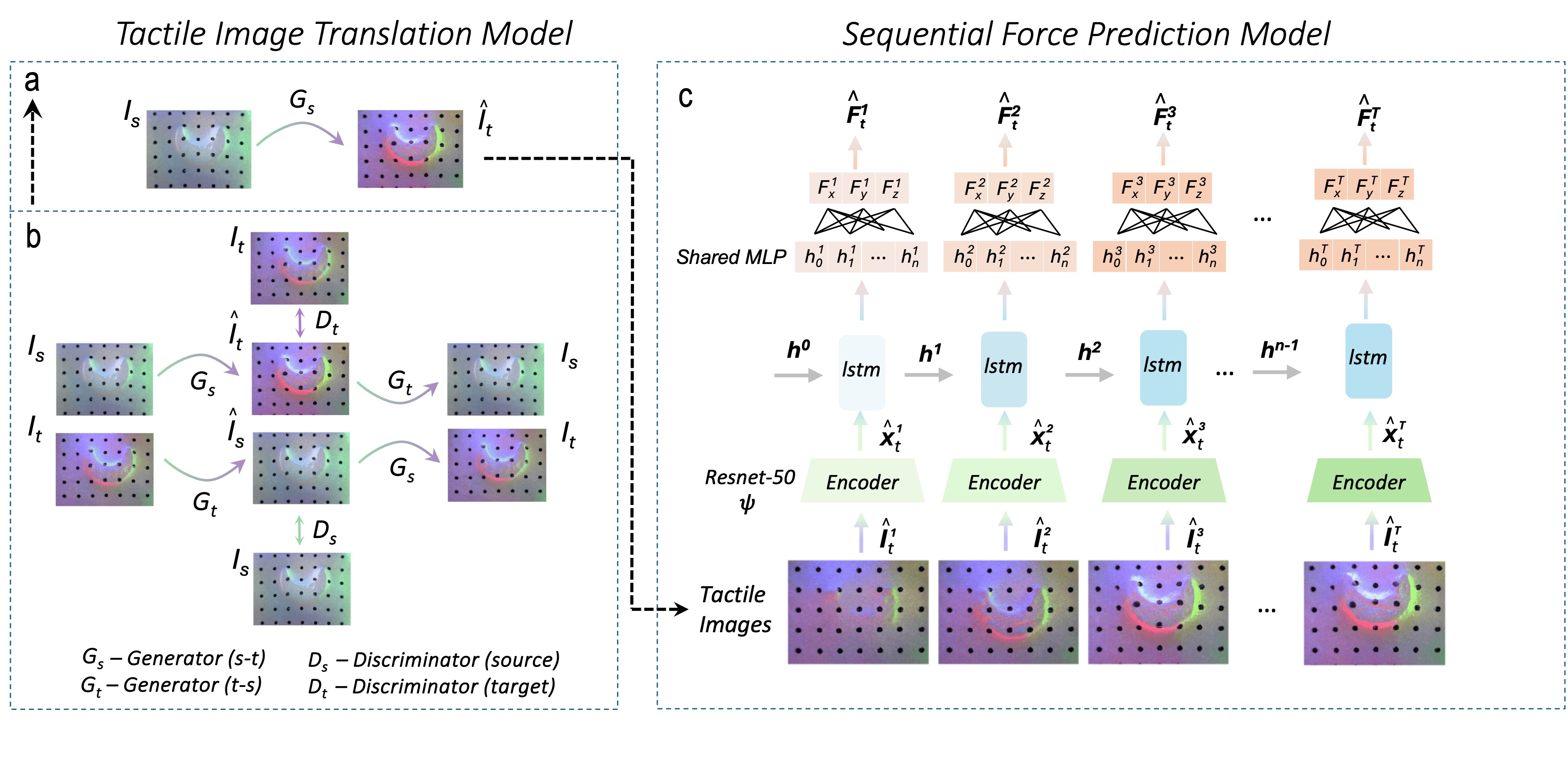}
\caption{Pipeline of the TransForce model. (a-b) Image translation process for (a) translating the tactile image $\mathbf{I}_s$ to $\hat{\mathbf{I}}_t$ by (b) training a generative model with generator $G_s$ mapping images from $\mathcal{S} \rightarrow \mathcal{T}$ and generator $G_t$ mapping images from $\mathcal{T} \rightarrow \mathcal{S}$. The discriminator $D_t$ aims to discriminate $\hat{\mathbf{I}}_t$ from $\mathbf{I}_t$ while $D_s$  aims to discriminate $\hat{\mathbf{I}}_s$ from  $\mathbf{I}_s$. (c) Sequential force prediction model. The model is trained with sequential generated images $\hat{\mathbf{I}}^1_t \sim \hat{\mathbf{I}}^T_t$ and force labels $\mathbf{F}^1_s \sim \mathbf{F}^T_s$, while predicts forces $\hat{\mathbf{F}}^1_t \sim \hat{\mathbf{F}}^T_t$ by taking $\mathbf{I}^1_t \sim \mathbf{I}^T_t$ as input. }
\label{fig2}
\end{figure*}

\section{RELATED WORK}
\label{sec:relatedworks}
\subsection{Vision-based Tactile Sensors}\label{R1}

Vision-based tactile sensors (VBTSs)~\cite{GelSight_review} are distinguished by their ability to capture high-resolution images, unlike traditional signal-based tactile sensors~\cite{chen2022laser}. Over the past decades, researchers have developed various structural designs for VBTSs to support diverse tasks in unstructured environments, including flat-shaped~\cite{GelSight_slip,lambeta2020digit}, fingertip-shaped~\cite{insight,geltip,Tactip}, edge-shaped~\cite{gelwedge}, and bubble-shaped~\cite{peng20243d} sensors. Innovations in light sources, reflective layers, and cameras aim to achieve specific functionalities, such as marker-switchable setups for versatile robotic tasks~\cite{ou2024marker}, colored illumination for depth reconstruction~\cite{lin20239dtact}, integrated motors for rotational capability~\cite{jiang2024rotipbot}, and the use of event cameras for high-frequency processing~\cite{funk2024evetac}. While these advancements enrich the VBTS community by providing a variety of options for robotic applications, they also introduce domain gaps among different VBTS types. This necessitates re-collection of data and re-training of neural networks for each sensors in specific tasks.

\subsection{Force Sensing of Vision-based Tactile Sensors}\label{R2}

Force prediction using VBTSs can be categorized into two primary approaches: physical-model-based methods and neural-network-based methods. In physical-model-based methods, force mapping is approached as an inverse problem of deformation estimation. This requires the establishment of a stiffness matrix~\cite{zhang2022tac3d} or a mapping function~\cite{li2024biotactip} specific to each sensor. Force mapping relationships can be based on factors such as light intensity~\cite{fu2024eltac}, marker displacement~\cite{zhang2023improving, fernandez2021visiflex}, and stresses~\cite{sui2021incipient}. The Finite Element Method (FEM) has also been shown to effectively address such inverse problems~\cite{ma2019dense, taylor2022gelslim}. However, these methods often assume soft elastomers with linear deformation behavior under small force and may perform poorly under large deformation~\cite{fernandez2021visiflex, hyperelastic}. In contrast, neural-network-based methods utilize data-driven approaches and demonstrate impressive performance on handling nonlinearity across wide force range~\cite{insight, do2023densetact, lin20239dtact}. However, these methods are data-hungry and exhibit poor generalization when sensor components change. Consequently, re-collecting paired tactile image-force data and re-training force prediction models is still necessary. Transferring existing image-force pairing to new sensors and training force prediction models with high accuracy remain significant challenges.

\subsection{Image-to-Image Translation for Tactile Images}\label{R3}

Generative Adversarial Networks (GANs)~\cite{goodfellow2014generative} have become widely used for image generation, achieving impressive results in style transfer. By optimizing an adversarial loss, GANs enable the generator to translate input images to target images that are indistinguishable from real images by the discriminator. In robotics, GANs are primarily employed for sim-to-real or real-to-sim transfer, adapting skills learned in simulated environments to real-world applications~\cite{kim2023marker, rao2020rl}. For instance, Quan et al.~\cite{luu2023simulation} utilized GANs to convert real observations from TacLink into simulated tactile images for contact shape reconstruction. Fan et al.~\cite{fan2024vitactip} trained two GANs to address marker removal and ambient light correction for modality conversion. CycleGAN ~\cite{zhu2017unpaired} extends GAN's capabilities by introducing a cycle consistency loss, which removes the need for paired images. Chen et al.~\cite{chen2022bidirectional} proposed a bidirectional sim-real transfer technique using CycleGAN for tactile depth construction and classification, while Jing et al.~\cite{jing2023unsupervised} enhanced image translation performance by integrating attention mechanisms. Although these methods excel in respective tasks, they are generally tailored towards sim-to-real translation or classification tasks \cite{jianu2022reducing}. There remains a need for an image translation framework to address regression challenges, such as the transferable force prediction for VBTSs, which is the focus of this work.

\section{METHODOLOGY}
\label{sec:methodology}
\subsection{Problem Definition}\label{Problem Definition}

In this problem, we are given two sets of tactile data from two VBTSs shown in Fig. \ref{fig1}: one that has pairs of tactile images and forces $\{\mathbf{I}^i_s, \mathbf{F}^i_s\}_{i=1}^{n_s}$, named source domain $\mathcal{S}$, and the other one that only has tactile images $\{\mathbf{I}^i_t\}_{i=1}^{n_t}$, named target domain $\mathcal{T}$. We assume that both sensors are the same type of VBTSs, such as GelSight \cite{GelSight_review}, GelTip \cite{geltip}, et.al, and comprise soft elastomer fabricated under identical conditions. As such, when identical forces \(\mathbf{F}_s = \mathbf{F}_t\) are applied to the above two sensors, the deformation of the soft elastomers should be the same. Nevertheless, the tactile images \(\mathbf{I}_s\) and \(\mathbf{I}_t\) differ due to domain gaps, i.e., the variations in illumination colors and marker patterns. Therefore, the goal is to align the image styles of \(\mathbf{I}_s\) and \(\mathbf{I}_t\) so that we can map \(\mathbf{I}_s\) to \(\mathbf{\hat{I}}_t\), which shares the same style with $\mathbf{I}_t$. To this end, we can use the image-force pairs \(\{\mathbf{\hat{I}}^i_t, \mathbf{F}^i_s\}_{i=1}^{n_s}\) derived from \(\mathcal{S}\) to train a mapping function \(\hat{\phi}\). This function \(\hat{\phi}\) is able to estimate forces \(\{\mathbf{F}^i_t\}\) in \(\mathcal{T}\) using the tactile images \(\{\mathbf{I}^i_t\}_{i=1}^{n_t}\).

Specifically, the data distributions can be denoted as \(\mathbf{I}_s \sim p(\mathbf{I}_s)\) for \(\mathcal{S}\) and \(\mathbf{I}_t \sim p(\mathbf{I}_t)\) for \(\mathcal{T}\). The ground truth for the applied forces in \(\mathcal{S}\) is represented as \(\mathbf{F}^i_s = (F^i_{x}, F^i_{y}, F^i_{z})\), encompassing the normal force \(F^i_{z}\) and the shear forces \(F^i_{x}\) and \(F^i_{y}\). Our first objective is to train a generator \(G_s\) that maps \(\mathbf{I}_s\) to \(\mathbf{\hat{I}}_t\), defined as \(G_s:\mathbf{I}_s \rightarrow \mathbf{\hat{I}}_t\), where \(G_s(\mathbf{I}_s) = \mathbf{\hat{I}_t}\). This generator aims to align the distribution of the generated tactile images \(p(\mathbf{\hat{I}_t})\) with the distribution of the target domain \(p(\mathbf{I}_t)\). To achieve this, a discriminator \(D_T\) is introduced to differentiate between the real tactile images \(\{\mathbf{I}_t\}\) and the generated tactile images \(\{\mathbf{\hat{I}_t}\}\). Once the generator is trained effectively, \(p(\mathbf{\hat{I}_t})\) should approximate \(p(\mathbf{I}_t)\). We can then utilize the generated sequential tactile images \(\{\mathbf{\hat{I}}^i_t\}_{i=1}^T\) as input and \(\{\mathbf{F}^i_s\}_{i=1}^T\) as the ground truth to train a regressor $\hat{\phi}$. The regressor \(\hat{\phi}:\mathbf{I} \rightarrow \mathbf{F}\) is able to predict \(\mathbf{\hat{F}}_t\) with $\mathbf{I}_t$, which approximates $\mathbf{F}_t$ as close as possible:

\begin{equation}
\arg \min_{\hat{\phi}} \mathbb{E}_{(\mathbf{I}_t, \mathbf{F}_t)} \left \lVert \hat{\phi}(\mathbf{I}_t) - \mathbf{F}_t \right \lVert_1
\end{equation}

\subsection{TransForce Model}\label{TransForce Model}

As illustrated in Fig. \ref{fig2}, our model consists of two components: a generative model for translating tactile images from $\mathbf{I}_s$ to $\mathbf{\hat{I}}_t$, and a force prediction model for predicting forces $\{\mathbf{\hat{F}}^i_t\}^{T}_{i=1}$ trained with generated sequential tactile images $\{\mathbf{\hat{I}}^i_t\}^{T}_{i=1}$ and force labels $\{\mathbf{\hat{F}}^i_s\}^{T}_{i=1}$. To train the generative model with unpaired tactile images from \(\mathcal{S}\) and  \(\mathcal{T}\), we employ CycleGAN \cite{zhu2017unpaired}. Our objective function for the generative model includes three key losses: the adversarial loss \(\mathcal{L}_{adv}\), the cycle consistency loss \(\mathcal{L}_{cyc}\), and the identity loss \(\mathcal{L}_{idt}\). These losses guide the generative model to produce desired translation between twodomains:

\noindent
\textbf{Adversarial Loss.} The adversarial loss $\mathcal{L}_{adv}$ aims at training two pairs of generator-discriminator, i.e.,  $G_s\sim D_t$ and $G_t\sim D_s$, to approximate the distribution of $p(\mathbf{\hat{I}}_t)$ to $p(\mathbf{I}_t)$ and $p(\mathbf{\hat{I}}_s)$ to $p(\mathbf{I}_s)$ shown in Fig. \ref{fig2}b. Specifically, for mapping $G_s:\mathbf{I}_s \rightarrow \mathbf{\hat{I}}_t$, the adversarial loss is defined as:
\begin{equation}
    \begin{aligned}
\min_{G_s}\max_{D_t}\mathcal{L}_{adv}(G_s,&D_t)  = \mathbb{E}_{\mathbf{I}_t\sim p(\mathbf{I}_t)}[\text{log}D_t(\mathbf{I}_t)] \\
                        & + \mathbb{E}_{\mathbf{I}_s\sim p(\mathbf{I}_s)}[\text{log}(1-D_t(G_s(\mathbf{I}_s))]
    \end{aligned}
\end{equation}
where the generator $G_s$ aims at generating $\mathbf{\hat{I}_t}$ and minimizing the \(\mathcal{L}_{adv}\), while $D_t$ is leveraged to discriminate $\mathbf{\hat{I}_t}$ and $\mathbf{I}_t$ as well as maximize \(\mathcal{L}_{adv}\). The same applies to $\mathbf{I}_t \rightarrow \mathbf{\hat{I}}_s$, which shares the same style of $\mathbf{I}_s$.

\noindent
\textbf{Cycle Consistency Loss.} The cycle consistency loss \(\mathcal{L}_{cyc}\) prevents the generation of artifacts by translating \(\mathbf{\hat{I}}_t\) back to \(\mathbf{I}_s\) and ensures the learned generator preserve essential image contents, such as indenters' shape and contact depth when from $\mathbf{I}_s$ to $\hat{\mathbf{I}}_t$:
\begin{equation}
    \begin{aligned}
    \min \mathcal{L}_{cyc}(G_s,G_t) & = \mathbb{E}_{\mathbf{I}_s\sim p(\mathbf{I}_s)}[\left \lVert G_t(G_s(\mathbf{I}_s)) - \mathbf{I}_s \right \lVert_1] \\
    & + \mathbb{E}_{\mathbf{I}_t\sim p(\mathbf{I}_t)}[\left \lVert G_s(G_t(\mathbf{I}_t)) - \mathbf{I}_t \right \lVert_1]
    \end{aligned}
\end{equation}
where the forward process $\mathbf{I}_s \rightarrow G_s(\mathbf{I}_s) \rightarrow G_t(G_s(\mathbf{I}_s))\rightarrow \mathbf{I}_s$ brings input tactile image $\mathbf{I}_s$ back to identical images in the end, and vice versa for the backward process $\mathbf{I}_t \rightarrow G_t(\mathbf{I}_t) \rightarrow G_s(G_t(\mathbf{I}_t)) \rightarrow \mathbf{I}_t$.

\noindent
\textbf{Identity Loss.} The identity loss \(\mathcal{L}_{idt}\) encourages bidirectional mapping to preserve the illumination colors when the input tactile image is already from the target domain.
\begin{equation}
    \begin{aligned}
    \min \mathcal{L}_{idt}(G_s,G_t) & = \mathbb{E}_{\mathbf{I}_s\sim p(\mathbf{I}_s)}[\left \lVert G_t(\mathbf{I}_s)) - \mathbf{I}_s \right \lVert_1] \\
    & + \mathbb{E}_{\mathbf{I}_t\sim p(\mathbf{I}_t)}[\left \lVert G_s(\mathbf{I}_t) - \mathbf{I}_t \right \lVert_1]
    \end{aligned}
\end{equation}

\noindent
\textbf{Objective for Image Translation.} The overall objective for the generative model combines above three losses, which are then weighted by $\lambda_{adv}$, $\lambda_{cyc}$, and $\lambda_{idt}$: 
\begin{equation}
\arg \min \lambda_{adv}\mathcal{L}_{adv} + \lambda_{cyc}\mathcal{L}_{cyc} + \lambda_{idt}\mathcal{L}_{idt}
\end{equation}

\noindent
\textbf{Objective for Force Prediction.} Upon completing image generation, a recurrent force prediction model is trained as depicted in Fig.~\ref{fig2}c. This model takes a sequence of tactile images \(\{\mathbf{\hat{I}}^i_t\}_{i=1}^T\) as input and uses corresponding forces \(\{\mathbf{F}^i_s\}_{i=1}^T\) as ground truth. Unlike training with single image in \cite{insight,chen2024deep,lin20239dtact}, this recurrent model memorises temporal information regarding elastomer's deformation, thereby promising in enhancing performance in shear force prediction. The model aims to minimize the $L_1$ distance between the predicted forces \(\hat{\mathbf{F}}_t\) and the ground truth \(\mathbf{F}_s\) across the entire sequence with an exponentially increasing weight $\gamma$ \cite{teed2020raft}:

\begin{equation}\label{equ6}
\min \mathcal{L}_{reg} = \sum_{i=1}^T\gamma^{T-i}\left\lVert \mathbf{F}^i_s-\mathbf{\hat{F}}^i_t\right\lVert_1
\end{equation}

\section{DATA COLLECTION AND IMPLEMENTATION}\label{Sec:IV}

\subsection{Data Collection}\label{dataCollect}

\begin{figure}[tbp]
\centering
\includegraphics[width=0.35\textwidth]{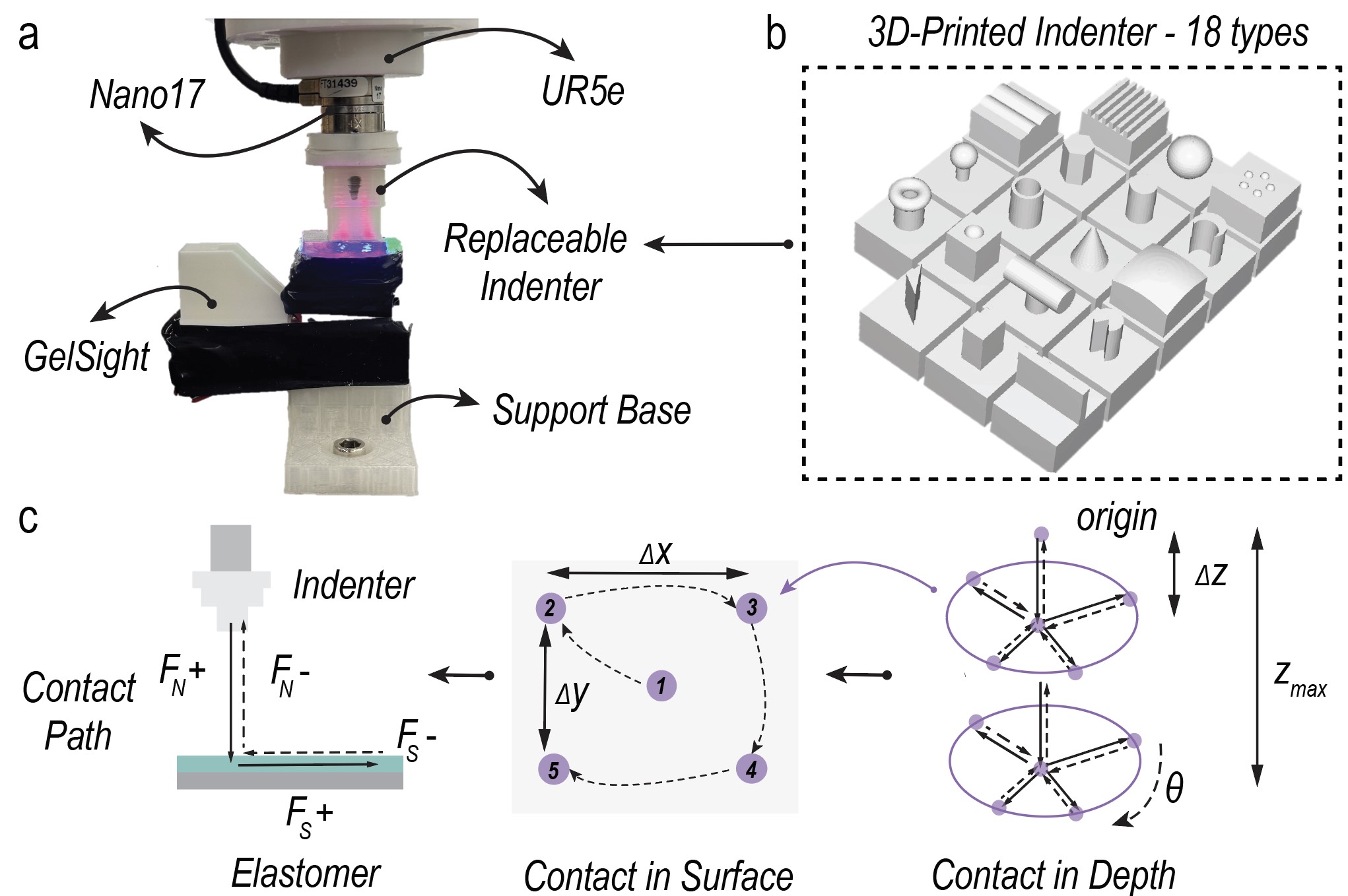}
\caption{(a) Real-world setup for data collection (b) 3D-printed indenters. (c) Contact path for applying normal force and shear force.}
\label{Fig3}
\end{figure}

As illustrated in Fig. \ref{Fig3}a, the real-world setup for collecting tactile image and force pairs comprises five main components: UR5e robot arm, Nano17 force/torque sensor, 3D-printed indenters \cite{chen2022bidirectional}, GelSight sensors \cite{GelSight_review}, and support base.  Before data collection, we fabricate two GelSight sensors using the same silicone elastomer (XP-565, ratio A:B = 1:15, size 10$\times$8 mm$^2$, thickness 3 mm), but with different marker patterns and illumination colors. Additionally, to test the generalization ability of our model, we print 18 different types of indenters with various shapes, roughness levels, and aspect ratios, as shown in Fig. \ref{Fig3}b. The contact paths for applying normal forces \(F_N\) and shear forces \(F_S\) are depicted in Fig. \ref{Fig3}c. Unlike our previous work~\cite{chen2024deep}, we employed five contact points on the surface due to a larger size of 3D-printed indenter, which adequately contacts the entire surface of soft elastomer. The contact motion was divided into four stages: downward movement, horizontal movement, inverse horizontal movement, and upward movement. This approach ensures that the collected data encompasses both of the increase process (\(\mathbf{F}_N^+, \mathbf{F}_S^+\)) and decrease process (\(\mathbf{F}_N^-, \mathbf{F}_S^-\)) of contact forces. Images and forces were synchronized and collected in real-time at 30 Hz, resulting in a total of 5$\times$4$\times$12$=$240 target points with varying angles, depths, and locations. Specifically, the moving distances \(\Delta x\) and \(\Delta y\) were 6 mm and 4 mm, the depth step \(\Delta Z\) was 0.2 mm with a maximum depth \(Z\) of 0.8 mm, and the moving angle \(\theta\) was \(30^\circ\). For each contact point, the number of collected frames varied based on the robot arm's speed and contact displacement. In our training phase, we randomly sampled eight image-force pairs from the sequential tactile images at each contact point for four times, resulting in a total of 18$\times$5$\times$4$\times$12$\times$4$=$17,280 tactile videos per sensor. The normal force range was from -16 N to 0 N, while the shear force range was from -6 N to 6 N.

\subsection{Implementation \& Evaluation Metrics}\label{iem}

For training the generative model, we adopt the architecture of CycleGAN~\cite{zhu2017unpaired}. The indenters into two groups with a 12:6 ratio: $seen$ and $unseen$. Specifically, the $unseen$ group includes $sphere$-$s$, $triangle$, $pacman$, $wave$, $torus$ and $cone$ indenters, while the $seen$ group comprises the remaining 12 types. The collected tactile images $\mathbf{I}_s$ and $\mathbf{I}_t$ are visualized in Fig.~\ref{fig4}. During training, we set $\lambda_{adv}=1$, $\lambda_{cyc}=10$, and $\lambda_{idt}=0.5$. We trained the model with 30 epochs with a total of 92,160 images per epoch. For the force prediction model, we use a shared ResNet-50 as the feature encoder to transform input sequential images into feature embeddings $\hat{\mathbf{X}}^T_t$ with a dimension of 64. An LSTM layer then processes $\hat{\mathbf{X}}^T_t$ to produce hidden states $\hat{\mathbf{h}}^T$ with the same dimension. A shared multilayer peceptron (MLP) with sigmoid activation function maps the hidden state to forces $\hat{\mathbf{F}}^T_t$. We set $\gamma=0.8$ in Eq.~\ref{equ6} to sum the $L_1$ distance. In Sec.~\ref{Force Prediction Result}, we include a control group named $nolstm$, where the output of the encoder is directly connected to the fully connected layer, and both models are trained for 20 epochs. To evaluate the impact of RGB images, marker-based images, and their combination (RGB$+$marker) on force prediction performance, we use the TELEA inpainting method to extract RGB images and morphology operations to obtain marker images from $\hat{\mathbf{I}}^T_t$. The learning rates are set to 0.0002 for image translation and 0.1 for force prediction.

\section{EXPERIMENTAL RESULTS}\label{experiment}
\label{sec:experiment}

In this section, we first evaluate the image translation performance on the $seen$ group and $unseen$ group. Next, we present quantitative results for force prediction performance using the sequential model, comparing it with the source-only method and the TransForce model without LSTM. Finally, we analyze the impact of RGB images, marker-based images, and their combinations on force prediction performance. We use Frechet Inception Distance (FID) and Kernel Inception Distance (KID) to evaluate the image translation performance. The Mean Absolute Error (MAE) is used to quantify the force prediction errors while the coefficient of determination $R^2$ is leveraged to evaluate regression performance.

\subsection{Image Translation Result}\label{i2i result}

\begin{figure}[htbp]
\centering
\includegraphics[width=0.42\textwidth]{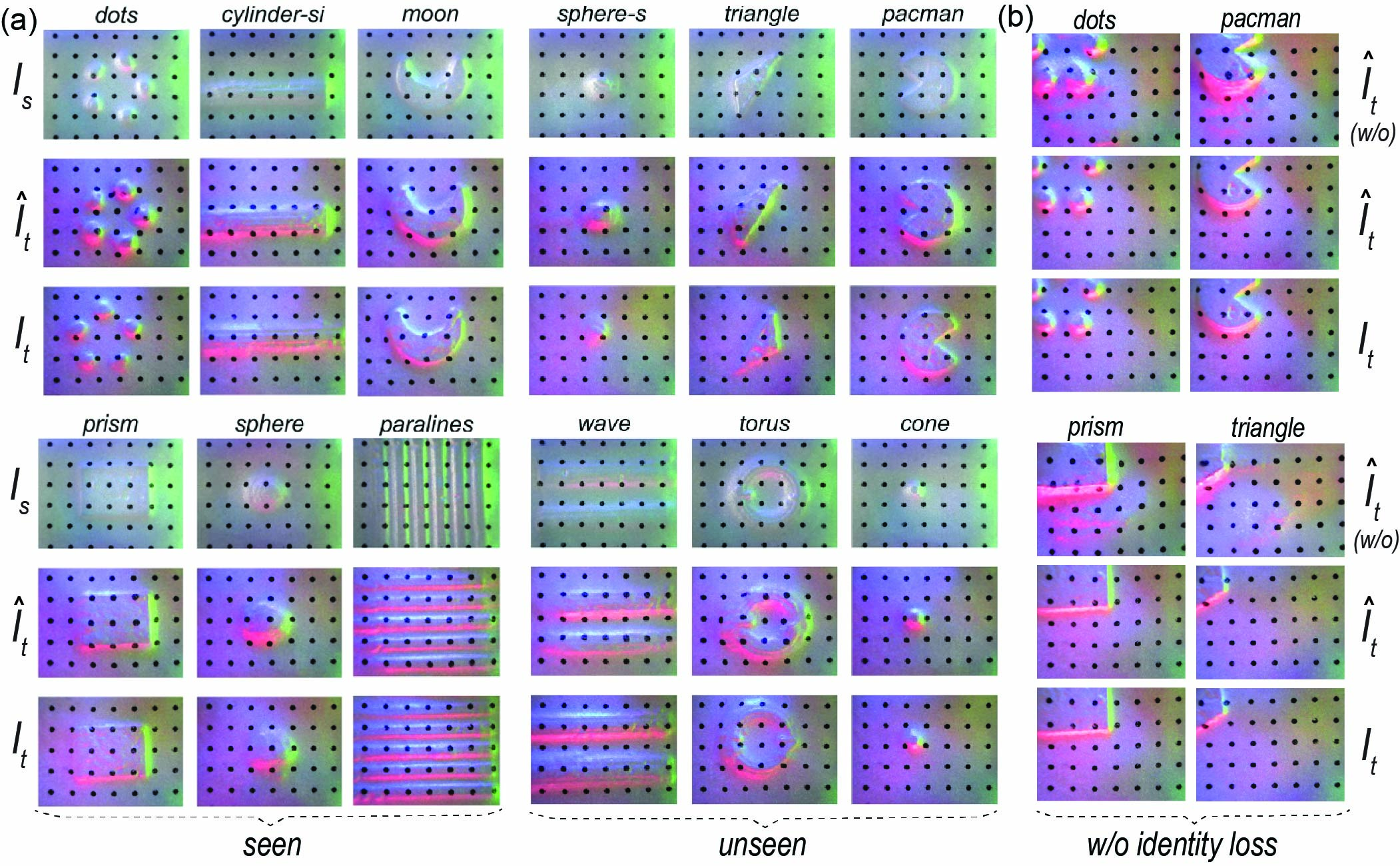}
\caption{(a-b) Visualization of tactile image translation with selected 6 types of 3D-printed indenters from $seen$ group and $unseen$ group. (b) Image translation results for models with and without (w/o) identity loss.}
\label{fig4}
\end{figure}

As discussed in Sec. \ref{Problem Definition}, the goal of image translation in this task is to generate tactile images $\mathbf{\hat{I}_t}$ that closely resemble $\mathbf{I}_t$ in terms of illumination colors and marker patterns, while preserving the indenters' shape, position, and elastomer deformation from $\mathbf{I}_s$. Fig. \ref{fig4}a and Table \ref{tab:Image Translation} demonstrates the image translation results. In $seen$ group, the generated tactile images $\mathbf{\hat{I}_t}$ closely match $\mathbf{I}_t$ in illumination colors and marker patterns, resulting approximately four times similarity on FID and KID when comparing $\mathbf{I}_s$ with $\mathbf{I}_t$ and comparing $\mathbf{I}_s$ with $\mathbf{\hat{I}_t}$. The tactile videos available in the supplementary material also show that frames of $\mathbf{\hat{I}_t}$ are synchronized with $\mathbf{I}_s$ and accurately reflect the contact pose and position. However, we identify a failure case with the $paralines$ indenter in the $seen$ group. This issue may stem from the large contact area of $paralines$, which covers nearly the entire field of view of our GelSight sensor. This extensive coverage likely cause the model to directly translate $\mathbf{I}_s$ to $\mathbf{\hat{I}_t}$ without preserving the specific pose of $paralines$. To test the generalizability, we evaluate the model on the $unseen$ group shown in Fig. \ref{fig4}a, which includes 6 indenters with varying aspect ratios and shapes. The results demonstrate that the model effectively translates marker patterns and illumination colors, with particularly good performance for indenters $triangle$, $pacman$, and $wave$, which maintain similar contact poses to $\mathbf{I}_s$. The values for FID and KID are nearly the same against $seen$ group. Additionally, we tested the model without the identity loss component. The results, shown in Fig. \ref{fig4}b, reveal that omitting identity loss leads to stray light and extra textures near the contact positions. This highlights the importance of the identity loss in preserving the fidelity of the illumination colors during tactile image translation.

\begin{table}[htbp]
\centering
\caption{Image Translation Evaluation}
\label{tab:Image Translation}
\resizebox{0.4\textwidth}{!}{
\begin{tabular}{|c|cc|cc|cc|l}
\cline{1-7}
\multirow{2}{*}{group} & \multicolumn{2}{c|}{$\mathbf{I}_s$ - $\mathbf{I}_t$}   & \multicolumn{2}{c|}{$\mathbf{I}_s$ - $\hat{\mathbf{I}}_t$} & \multicolumn{2}{c|}{$\mathbf{I}_t$ - $\hat{\mathbf{I}}_t$} & \multicolumn{1}{c}{} \\ \cline{2-7}
                       & \multicolumn{1}{c|}{FID$\downarrow$} & KID$\downarrow$ & \multicolumn{1}{c|}{FID$\downarrow$}   & KID$\downarrow$   & \multicolumn{1}{c|}{FID$\downarrow$}   & KID$\downarrow$   &                      \\ \cline{1-7}
seen                   & \multicolumn{1}{c|}{81.1}            & 0.126           & \multicolumn{1}{c|}{85.4}              & 0.133             & \multicolumn{1}{c|}{\textbf{18.3}}              & \textbf{0.019}             &                      \\ \cline{1-7}
unseen                 & \multicolumn{1}{c|}{87.9}            & 0.134           & \multicolumn{1}{c|}{92.4}              & 0.144             & \multicolumn{1}{c|}{\textbf{21.3}}              & \textbf{0.021}            &                      \\ \cline{1-7}
\end{tabular}%
}
\end{table}

\subsection{Force Prediction Result}\label{Force Prediction Result}

Compared to our previous work \cite{chen2024deep}, which was limited in force range and indenter variety, we have significantly expanded the scope in this study. Given its subpar performance of the DAR method in shear force prediction, we do not include it in our comparison. Instead, we directly compare our approach with the TransForce model using a source-only method with LSTM and a TransForce model without LSTM, where the source-only method represents training the model on $\mathbf{I}_s$ and directly predicting forces using $\mathbf{I}_t$. We eliminate use of $paralines$ of $seen$ group in the training phase as the reason shown in Sec. \ref{i2i result}.

\begin{figure*}[htbp]
\centering
\includegraphics[width=0.83\textwidth]{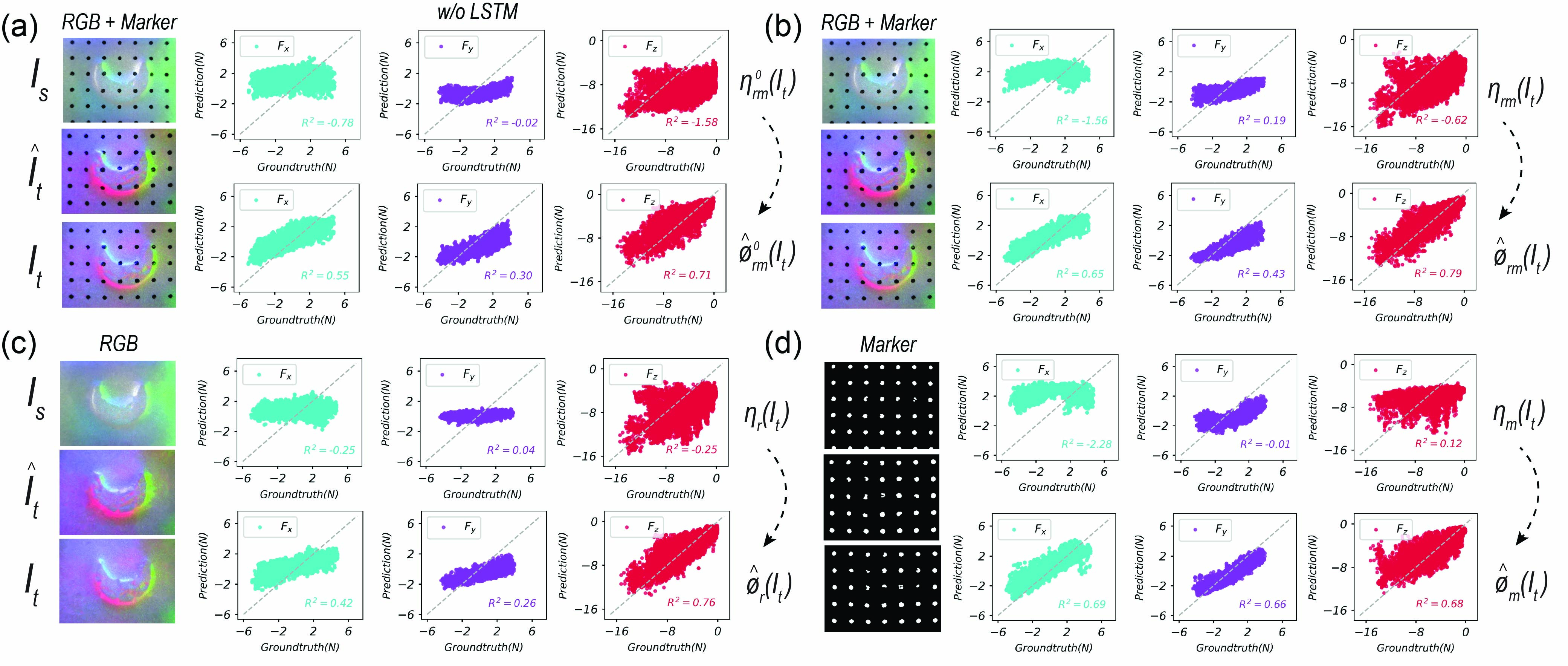}
\caption{Force prediction performance using source-only method ($top$ row in each panel, denoted as $\eta(\cdot)$) and TransForce model ($bottom$ row, denoted as $\phi(\cdot)$) with three types of tactile images, including (a-b) RGB images with markers $rm$, (c) RGB images without markers $r$, (d) marker-only tactile images $m$. Note that (a) is tested with the model without LSTM (with supersript $o$) while (b-d) are with LSTM.}
\label{fig5}
\end{figure*}

\begin{figure}[htbp]
\centering
\includegraphics[width=0.43\textwidth]{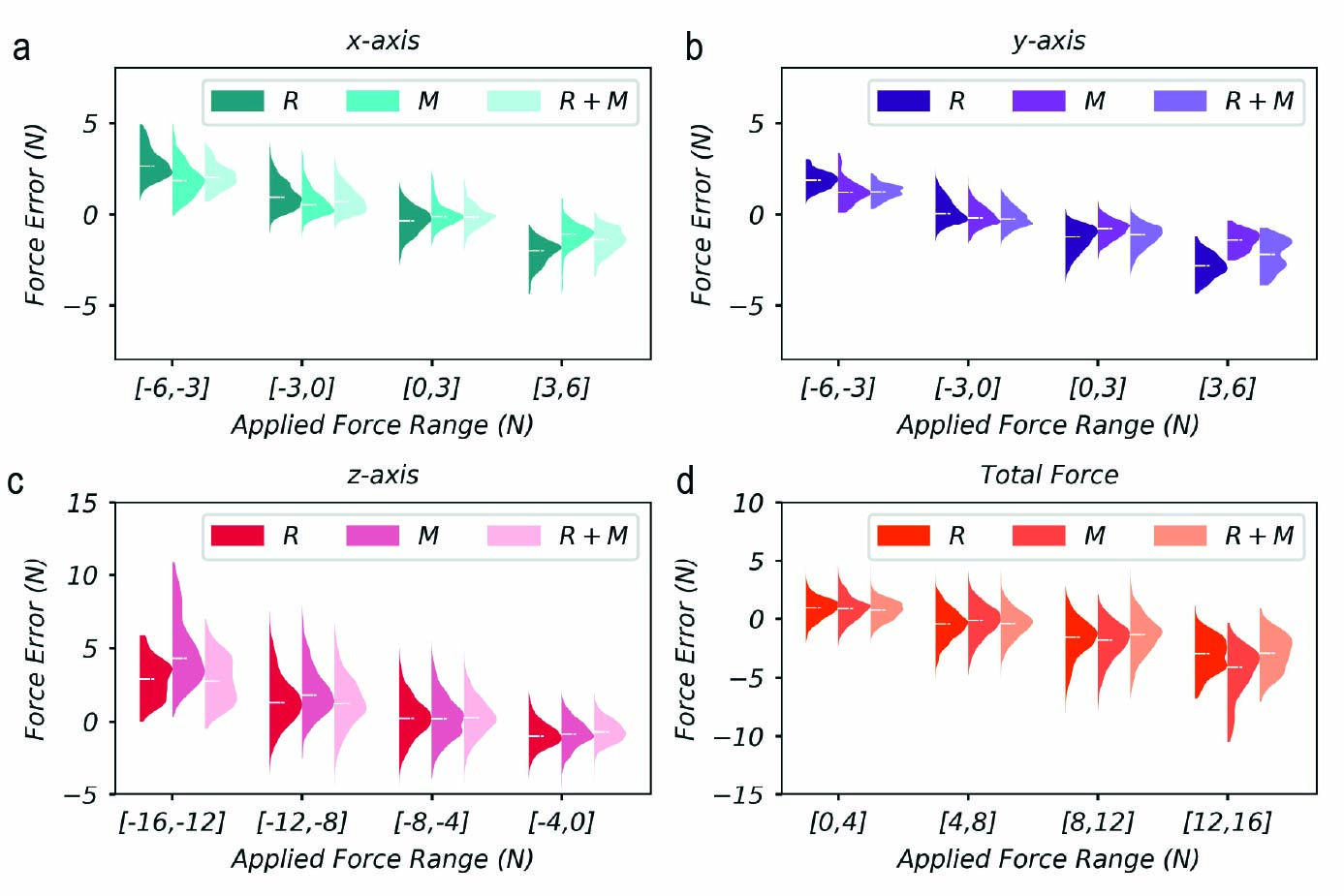}
\caption{Force prediction performance for TransForce model with LSTM in $seen$ group by taking RGB images ($R$), marker images ($M$), and their combinations ($R+M$) as inputs. The white lines denote median values.}
\label{fig6}
\end{figure}

Fig. \ref{fig5} illustrates the force prediction performance for the $seen$ group. When using the source-only method, whether with or without the LSTM module ($\eta^0_{rm}(\mathbf{I}_t)$ and $\eta_{rm}(\mathbf{I}_t)$), the predicted forces exhibit significant errors, as indicated by negative $R^2$ values in both normal and shear directions. This demonstrates the inadequacy of the source-only approach in adapting to new sensors. In contrast, the TransForce model shows marked improvements. As shown in Fig. \ref{fig5}a, even without using LSTM, the $R^2$ values forces in normal direction improve from -1.58 to 0.71, while the shear forces improve from -0.78 to 0.55 in $x$-axis and from -0.02 to 0.30 in $y$-axis. When trained with sequential images, as depicted in Fig. \ref{fig5}b, the $R^2$ values further improve to 0.65 for shear forces in $x$-axis ($F_x$), and 0.43 in the $y$-axis ($F_y$). Additionally, in normal force direction, $R^2$ increases to 0.79. Comparing these results with the source-only method and our previous work \cite{chen2024deep}, the $R^2$ values have been improved significantly in shear direction. This underscores the effectiveness of the TransForce model. For the predicted force accuracy of $\hat{\phi}_{rm}(\mathbf{I}_t)$, the error in shear direction is 0.771 N (6.4\% of the full range) in the $x$-axis and 0.899 N (7.5\%) in the $y$-axis. For the normal direction, the force prediction error is 1.112 N (6.9\%). These results validate the efficacy of our TransForce model in predicting forces accurately, especially for shear forces, even with expanded force ranges and diverse indenter types.

\subsection{Influence of tactile modalities on force prediction}

Among various types of tactile images used in VBTSs, three main modalities are prevalent in real-world robot applications: pure RGB images without markers, pure marker-based images, and a combination of both, as illustrated in Fig. \ref{fig5}. This raises the question of which modality is most suitable for force prediction and whether RGB information, which is commonly used for classification tasks \cite{chen2022bidirectional, fan2024vitactip}, is also essential for this task. To answer this, we compared force prediction results using above three types of tactile images. Results from the $seen$ group and $unseen$ group are presented in Table \ref{tab:seen_force} and Table \ref{tab:unseen_force}. Fig. \ref{fig6} illustrates the half-violin distributions of force error within four force ranges by using Transforce model with LSTM in $seen$ group.

The results indicate that, regardless of trained models, pure marker-based images (denoted as $M$) consistently yield the best performance in terms of shear force prediction, particularly with a lowest average error of 0.695N (5.8\% in full work range) in $x$-axis, 0.701N (5.8\%) in $y$-axis, and 1.112N (6.9\%) in $z$-axis using TransForce model with LSTM. This suggests that marker information is more crucial for accurate shear force prediction than RGB information, as markers explicitly involve shear displacement that is vital for slip detection \cite{GelSight_slip, sui2021incipient, james2020biomimetic, li2024incipient}. On the other hand, these models fed with marker modality combined with RGB information ($R+M$) show performance superior to models with pure marker images in normal forces prediction. This is due to that the RGB information complements contact depth and shape information for better prediction of normal force. To further analyse Table \ref{tab:seen_force} and Table \ref{tab:unseen_force}, we can find that the TransForce model with LSTM achieves the best in total force prediction. This finding underscores the importance of marker information in shear force prediction and the LSTM module in enhancing the overall force prediction accuracy. 

\begin{table}[]
\centering
\caption{Force Prediction Performance in \textbf{Seen} Group \\
Note: $error$ is calculated with MAE within full force range}
\label{tab:seen_force}
\resizebox{0.48\textwidth}{!}{%
\begin{tabular}{|c|c|cc|cc|cc|c|}
\hline
\multirow{2}{*}{Method}                                                           & \multirow{2}{*}{Image Type} & \multicolumn{2}{c|}{$F_x$}                                & \multicolumn{2}{c|}{$F_y$}                                & \multicolumn{2}{c|}{$F_z$}                                & Total Force          \\ \cline{3-9} 
                                                                                  &                             & \multicolumn{1}{c|}{$error(N)\downarrow$} & $R^2\uparrow$ & \multicolumn{1}{c|}{$error(N)\downarrow$} & $R^2\uparrow$ & \multicolumn{1}{c|}{$error(N)\downarrow$} & $R^2\uparrow$ & $error(N)\downarrow$ \\ \hline
\multirow{3}{*}{\begin{tabular}[c]{@{}c@{}}Source-only\\ ($lstm$)\end{tabular}}   & $R$                         & \multicolumn{1}{c|}{1.528}                & -0.25         & \multicolumn{1}{c|}{1.14}                 & 0.04          & \multicolumn{1}{c|}{3.085}                & -0.25         & 2.869                \\
                                                                                  & $M$                         & \multicolumn{1}{c|}{2.811}                & -2.28         & \multicolumn{1}{c|}{1.267}                & -0.01         & \multicolumn{1}{c|}{2.521}                & -0.71         & 2.771                \\
                                                                                  & $R+M$                       & \multicolumn{1}{c|}{2.425}                & -1.56         & \multicolumn{1}{c|}{1.092}                & 0.19          & \multicolumn{1}{c|}{3.623}                & 0.12          & 3.606                \\ \hline
\multirow{3}{*}{\begin{tabular}[c]{@{}c@{}}TransForce\\ ($no lstm$)\end{tabular}} & $R$                         & \multicolumn{1}{c|}{1.034}                & 0.38          & \multicolumn{1}{c|}{0.953}                & 0.31          & \multicolumn{1}{c|}{1.42}                 & 0.68          & 1.452                \\
                                                                                  & $M$                         & \multicolumn{1}{c|}{\textbf{0.731}}       & \textbf{0.67} & \multicolumn{1}{c|}{\textbf{0.754}}       & \textbf{0.62} & \multicolumn{1}{c|}{1.462}                & 0.64          & 1.495                \\
                                                                                  & $R+M$                       & \multicolumn{1}{c|}{0.891}                & 0.55          & \multicolumn{1}{c|}{0.994}                & 0.3           & \multicolumn{1}{c|}{\textbf{1.312}}       & \textbf{0.71} & 1.424                \\ \hline
\multirow{3}{*}{\begin{tabular}[c]{@{}c@{}}TransForce\\ ($lstm$)\end{tabular}}    & $R$                         & \multicolumn{1}{c|}{1.023}                & 0.42          & \multicolumn{1}{c|}{1.014}                & 0.26          & \multicolumn{1}{c|}{1.221}                & 0.76          & 1.283                \\
                                                                                  & $M$                         & \multicolumn{1}{c|}{\textbf{0.695}}       & \textbf{0.69} & \multicolumn{1}{c|}{\textbf{0.701}}       & \textbf{0.66} & \multicolumn{1}{c|}{1.34}                 & 0.68          & 1.384                \\
                                                                                  & $R+M$                       & \multicolumn{1}{c|}{0.771}                & 0.65          & \multicolumn{1}{c|}{0.899}                & 0.43          & \multicolumn{1}{c|}{\textbf{1.112}}       & \textbf{0.79} & \textbf{1.197}       \\ \hline
\end{tabular}%
}
\end{table}

\begin{table}[]
\centering
\caption{Force Prediction Performance in \textbf{Unseen} Group}
\label{tab:unseen_force}
\resizebox{0.48\textwidth}{!}{%
\begin{tabular}{|c|c|cc|cc|cc|c|}
\hline
\multirow{2}{*}{Method}                                                           & \multirow{2}{*}{Image Type} & \multicolumn{2}{c|}{$F_x$}                                & \multicolumn{2}{c|}{$F_y$}                                & \multicolumn{2}{c|}{$F_z$}                                & Total Force          \\ \cline{3-9} 
                                                                                  &                             & \multicolumn{1}{c|}{$error(N)\downarrow$} & $R^2\uparrow$ & \multicolumn{1}{c|}{$error(N)\downarrow$} & $R^2\uparrow$ & \multicolumn{1}{c|}{$error(N)\downarrow$} & $R^2\uparrow$ & $error(N)\downarrow$ \\ \hline
\multirow{3}{*}{\begin{tabular}[c]{@{}c@{}}Source-only\\ ($lstm$)\end{tabular}}   & $R$                         & \multicolumn{1}{c|}{1.462}                & -0.36         & \multicolumn{1}{c|}{0.913}                & 0.01          & \multicolumn{1}{c|}{4.215}                & -0.32         & 4.142                \\
                                                                                  & $M$                         & \multicolumn{1}{c|}{2.826}                & -2.71         & \multicolumn{1}{c|}{1.339}                & -0.26         & \multicolumn{1}{c|}{3.619}                & -0.06         & 4.095                \\
                                                                                  & $R+M$                       & \multicolumn{1}{c|}{2.548}                & -2.06         & \multicolumn{1}{c|}{0.995}                & 0.07          & \multicolumn{1}{c|}{4.341}                & -0.33         & 4.591                \\ \hline
\multirow{3}{*}{\begin{tabular}[c]{@{}c@{}}TransForce\\ ($no lstm$)\end{tabular}} & $R$                         & \multicolumn{1}{c|}{0.867}                & 0.28          & \multicolumn{1}{c|}{0.846}                & 0.2           & \multicolumn{1}{c|}{1.964}                & 0.65          & 1.944                \\
                                                                                  & $M$                         & \multicolumn{1}{c|}{\textbf{0.534}}       & \textbf{0.71} & \multicolumn{1}{c|}{\textbf{0.837}}       & \textbf{0.47} & \multicolumn{1}{c|}{1.78}                 & 0.64          & 1.844                \\
                                                                                  & $R+M$                       & \multicolumn{1}{c|}{0.667}                & 0.51          & \multicolumn{1}{c|}{1.084}                & 0.1           & \multicolumn{1}{c|}{\textbf{1.762}}       & \textbf{0.67} & 1.869               \\ \hline
\multirow{3}{*}{\begin{tabular}[c]{@{}c@{}}TransForce\\ ($lstm$)\end{tabular}}    & $R$                         & \multicolumn{1}{c|}{0.824}                & 0.26          & \multicolumn{1}{c|}{0.944}                & 0.14          & \multicolumn{1}{c|}{1.756}                & 0.71          & 1.765                \\
                                                                                  & $M$                         & \multicolumn{1}{c|}{\textbf{0.506}}       & \textbf{0.71} & \multicolumn{1}{c|}{\textbf{0.738}}       & \textbf{0.52} & \multicolumn{1}{c|}{1.744}                & 0.63          & 1.790                \\
                                                                                  & $R+M$                       & \multicolumn{1}{c|}{0.635}                & 0.53          & \multicolumn{1}{c|}{1.031}                & 0.19          & \multicolumn{1}{c|}{\textbf{1.613}}       & \textbf{0.75} & \textbf{1.709}       \\ \hline
\end{tabular}%
}
\end{table}

\section{CONCLUSION}
\label{sec:conclusion}

In this study, we propose a novel transferable model to address the challenge of unsupervised force prediction for VBTSs. Our approach effectively performs tactile image translation, transforming images from existing sensors to new sensors for leveraging collected force labels. A force prediction model that operates on sequential tactile images demonstrates superior force prediction accuracy over models trained with single image in both normal direction and shear direction. Our results also indicate that pure marker-based tactile images are the most effective modality for shear force prediction, while RGB information complements its inferiority for normal force prediction. We believe this work points out a useful technique to the VBTSs community regarding fast, low-cost force calibration of tactile sensors by leveraging existing paired image-force data. Owing to the generalization performance of generative model, this method is versatile and applicable to various types of VBTSs. It is also promising to be beneficial to large-scale integration of VBTSs on humanoid robots or robot hands with the demand of force sensing in the future. 

\bibliographystyle{IEEEtran}
\bibliography{manuscript}
\end{document}